\documentclass[10pt,journal,compsoc]{IEEEtran}

\usepackage{times}
\usepackage{epsfig}
\usepackage{graphicx}
\usepackage{amsmath}
\usepackage{amssymb}
\usepackage{float}

\DeclareMathAlphabet{\mathcal}{OMS}{cmsy}{m}{n}

\def\argmin{\mathop{\rm argmin}}

\newcommand{\inner}[2]{\left\langle #1,#2 \right\rangle}

\newcommand{\real}{\ensuremath{\mathbb{R}}}

\newcommand{\ltwo}{\ensuremath{\mathbb{L}^2}}

\newcommand{\norm}[1]{\left\lVert #1 \right\rVert}

\newcommand{\Rn}[1]{\real^{#1}}

\newcommand{\fn}{\mathcal{F}}

\newcommand{\trace}[1]{\textrm{trace}\left(#1\right)}
\newcommand{\KL}[2]{D_{KL}\left(#1 \; || #2\right)}
\newcommand{\Lcal}{\mathcal{L}}

\ifCLASSOPTIONcompsoc
  \usepackage[nocompress]{cite}
\else
  \usepackage{cite}
\fi

\begin{document}

\title{SrvfNet: A Generative Network for Unsupervised \\ Multiple Diffeomorphic Shape  Alignment}

%
%
%
%

\author{Elvis~Nunez,~\IEEEmembership{Student Member,~IEEE,}
        Andrew~Lizarraga,
        and~Shantanu~H.~Joshi,~\IEEEmembership{Senior Member,~IEEE}
\IEEEcompsocitemizethanks{\IEEEcompsocthanksitem E.~Nunez is with the Department of Electrical and Computer Engineering, A.~Lizarraga, and S.~H.~Joshi are with the UCLA Brain Mapping Center, Department of Neurology and Bioengineering, UCLA, CA 90095.\protect\\
elvis.nunez@,andrewlizarraga@mednet,s.joshi@g.\{ucla.edu\}}
\thanks{Manuscript received April 27, 2021.}}%

%
%

\IEEEtitleabstractindextext{%
\begin{abstract}
 We present  SrvfNet, a generative deep learning framework for
    the joint multiple alignment of large collections of functional data comprising square-root velocity functions (SRVF) to their templates. Our proposed framework is fully unsupervised and 
    is capable of aligning to a predefined 
    template as well as jointly predicting an optimal template from data while simultaneously achieving alignment. 
    Our network is constructed as a generative encoder-decoder architecture 
    comprising fully-connected layers capable of
    producing a distribution space of the warping functions. 
    We demonstrate the strength
    of our framework by validating it on synthetic data
    as well as diffusion profiles from magnetic resonance imaging (MRI) data.
\end{abstract}

}

\maketitle

\IEEEdisplaynontitleabstractindextext

\IEEEpeerreviewmaketitle

\IEEEraisesectionheading{\section{Introduction}\label{sec:introduction}}

%
%
%
%


 


\ifCLASSOPTIONcaptionsoff
  \newpage
\fi


\IEEEPARstart{O}{wing} to technological advances in imaging and sensing, the  availability of commercial  wearable devices, biosensors, and continuous recording instruments at reduced cost, there is an abundance of feature-rich scalar data. Such data arise from diverse applications and processes that produce continuous functions such as voltage, current from electronic meters, time-series data related to  electroencephalography or electrocardiography from medical applications, scalar  measures observed over spatial data from geological applications, or intensity-based features from spatial biological data. Recent advances in statistical shape and functional data analysis allow researchers to collectively study such data using geometric representations and structures~\cite{srivastava2016functional}. 
 
An important step in the statistical analysis of multiple signals or recordings is matching or aligning them across the population. Pairwise matching or alignment is convenient  for discriminative analysis such as classification and clustering, while the computation of the population mean is useful for inference. A typical approach for aligning multiple functional data  involves a two step process: i) compute a mean in the appropriate space equipped with a choice of a metric by minimizing the variance of the set of functions, often iteratively in case a closed form solution is not available, and ii) align or register all the elements in the dataset to this mean, thereby establishing a correspondence across the entire population. The procedure for the computation of the mean is  slow  and typically yields a local solution if gradient descent is used. Further, in the case of signals and time-series, if one seeks invariance to reparameterization (or arbitrary nonlinear order-preserving warpings of the domain of the function), then one needs to employ dynamic time warping (DTW) procedures~\cite{sakoe1978dynamic}. Although efficient algorithms exist, this process can quickly become computationally intensive for $T$-length data  $O(T^N)$ as the size of the data ($N$) increases. If DTW is applied at every iteration until the convergence of the mean, the computational cost can become prohibitively expensive for a large population ($N > 500K$). A slightly faster yet sub-optimal solution is to compute a Euclidean (extrinsic) average, project it back onto the space of functions and, using it as a template, compute dynamic time warps with respect to this template. However, such an average template may not be optimal with respect to the underlying metric, and because it was computed by ignoring the nonlinear domain warps, may not preserve important geometric features present in the data. 

In this paper, we propose a novel neural network architecture \emph{SrvfNet} inspired from variational auto-encoders to simultaneously predict multiple functional data alignments in addition to the population templates.

\subsection{Background and Related Work}
Several researchers have proposed ideas for machine-learning based estimation of nonlinear alignment. Here, we discuss the approaches related to our work. A framework based on Gaussian process-based non-parametric priors was proposed by Kazlauskaite et al.~\cite{kazlauskaite2018gaussian} to learn a latent model for functional alignments. A deep-learning based approach, the deep canonical time warping (DCTW) was used to perform joint temporal registration, while at the same time maximizing the joint correlation of multiple time-series~\cite{trigeorgis2017deep}. A sequence transformer network was proposed by Oh et al.~\cite{Oh_invariance:2018} to perform stretching, compression, shifting, and flipping of the time-series signals to incorporate invariant matching of signals from clinical data. The  approach by Lohit et al.~\cite{lohit_transformer} uses a temporal transformer network (TTN) to  learn the inter-signal warping functions while performing classification simultaneously. The TTN performs a joint discriminative alignment of time-series by decreasing the intra-class variability and increasing the inter-class separation without using a fixed template for each class. Abid et al.~\cite{Abid:18} proposed Auto-warp, which learns and optimizes a metric distance on the set of unlabeled time-series. This is achieved by pairwise alignment of signals without diffeomorphic constraints on the warping functions. Recently, Weber et al.~\cite{Weber:19} have proposed the learning of diffeomorphisms in an unsupervised manner. Their approach uses temporal transformer layers to perform joint alignment of time-series and includes a loss function that aims to minimize the empirical variance of warped signals while regularizing the network using continuous piecewise-affine (CPA) covariance matrices \cite{freifeld2015highly}. Koneripalli et al.~\cite{koneripalli2020rate} also use autoencoders and temporal warping layers for unsupervised learning of diffeomorphisms.
The reader is also referred to a previous approach by Nunez and Joshi~\cite{nunez_CVPRW:20} where a convolutional network is used for predicting warping functions when matching a pair of shapes.

\subsection{Contributions}
In this paper we develop a  deep neural network architecture, SrvfNet, that allows for unsupervised multiple  diffeomorphic alignment of functional data. We show that this generative-based architecture is capable of generating warping functions to warp data to a given template, and can simultaneously learn a suitable template when one is not provided.
Our method is unsupervised in the sense that only the population of functions we wish to align are input--that is, we 
do not require pre-computed warping functions nor a priori labels/targets.
Instead, our loss function aggregates an intrinsic geometric distance of the predicted alignments which is
then iteratively minimized to improve the predicted alignments. 
Further, a key distinction between our work and others is in our shape representation of functional data through the 
use of the Square-Root Velocity Function (SRVF) ~\cite{S.H.Joshi_etal_CVPR:07}.
Our network architecture is simple and consists only of fully-connected and regularization layers.
The design of our network is inspired by the variational autoencoder (VAE) \cite{kingma2013} in the sense that
inputs are encoded onto a low-dimensional latent space which then serves to model the conditional posterior
distribution of the warping functions. Contrary to the autoencoder
design of VAEs, we do not measure a reconstruction error, nor do we impose a distribution 
assumption on the posterior distribution of the inputs given the latent variables. 
VAEs are trained to maximize the evidence lower bound (ELBO) function. While our loss functions
partially build off of the ELBO function, we do not include the term that arises from the posterior
distribution assumption made by VAEs. Instead, one novelty of our approach is in the replacement
of this term with a geometric measure.

This generative approach 
distinguishes our work from recent developments such as that of \cite{Weber:19,koneripalli2020rate,nunez_CVPRW:20} where the produced 
outputs are deterministic. Our approach has the added benefit of providing a statistical summary of 
the intrinsic relationship between the class of training data and the provided or predicted template. 
More specifically, our framework differs from \cite{Weber:19} in the following ways: i) our 
architecture does not use any temporal transformer layers nor any CPA-based transformations 
\cite{freifeld2015highly}, instead our network consists of simple feed-forward fully-connected layers 
with basic batch normalization and dropout regularizations; ii) our loss functions do not promote 
smoothness in the diffeomorphisms through a CPA-based regularization penalty and instead do so 
by penalizing the gradients of the predicted warps; iii) we approach this problem through a geometric
lens and consider a representation of the data that encodes geometric variability rather than working
with the raw data--consequently, our loss functions aim to minimize an intrinsic distance over a 
Hilbert sphere; iv) our framework is generative and provides a distribution over possible diffeomorphisms. 
Different from \cite{koneripalli2020rate}, we use a variational-based autoencoder rather than a 
deterministic one. Moreover, we do not aim to learn network weights by minimizing a mean squared
reconstruction error, but rather minimize a geometric measure on the predicted alignments. To this end, our framework 
allows for direct penalties over learned diffeomorphisms to control for smoothing that in turn lead to more 
robust alignments. 
Our paper also differs from \cite{nunez_CVPRW:20} where they used a supervised
deep learning framework to predict warps between pairs of shapes. Unlike their approach, our network can 
not only predict pairwise warping between functions, but can also perform unsupervised template estimation of multiple functions taken together.
 
This paper is organized as follows. Section \ref{sec:methods} outlines the geometric representation
and matching problem we are trying to solve. This section also derives 
our loss functions for both the fixed template and template prediction settings as well as the
diffeomorphic constraint satisfaction layers built into our network. Section \ref{sec:network_design}
discusses the architecture and training details of our network. Section \ref{sec:experiments}
presents the results of several experiments performed for both fixed and predicted template problems.
In section \ref{sec:discussion} we discuss the results of our experiments and in section
\ref{sec:conclusion} we provide concluding remarks. 


\section{Methods}\label{sec:methods}
\subsection{Shape Representation and Preliminaries} \label{sec:shape_rec}

We first provide the reader a brief summary of the shape representation of curves that will be used through out this paper. We represent curves as one-dimensional parameterized
functions. These functions are assumed to be first differentiable, and belonging to the class of $\ltwo$ functions, denoted as $f \in
\ltwo([0,1],\real)$, where an entire of collection
of curves is denoted $\fn \equiv \left\{f_i \mid f_i \in \ltwo([0, 1], \real)\right\}$.
We define a template as a designated
fixed function $g \in \ltwo([0,1],\real)$, and we seek to align an entire
collection of curves, $\fn$, to the template curve.

When performing computations and shape comparisons of curves, we represent
them as SRVFs~\cite{S.H.Joshi_etal_CVPR:07,Joshi2007emmcvpr,srivastava_etal_PAMI:11}. Each $f_i$ is represented by the SRVF map
$f_i \mapsto q_i = \frac{\dot{f_i}}{\sqrt{\norm {{\dot{f_i}}}}}$, and the
space of such functions is
denoted by $\mathcal{S}$. Adding the constraint, $\int_{[0,1]} \inner{q(t)}{q(t)} dt = 1$ under the standard $\ltwo$ inner product ($\inner{\cdot}{\cdot}$) ensures
that our shape representations are invariant to translation and scaling. As defined,
the SRVFs are unit normalized, thus shape comparisons are equivalent to computing
geodesic distances between SRVF points on the unit Hilbert Sphere $\mathcal{S}$ in
$\ltwo([0,1],\real)$. We introduce an analogous notation for $\mathcal{S}$ where we designate a class of functions as $\mathcal{Q} \equiv \left\{q_i \mid q_i \in \mathcal{S}\right\}$.

When warping a function $f$ to a specified template $g$, we solve the
following optimization problem in terms of the Fisher-Rao metric via dynamic programming~\cite{srivastava_etal_CVPR:07}
\begin{equation}
\argmin_{\gamma}{{\norm{q_f - \sqrt{\dot{\gamma}}(q_g \circ \gamma)}}}^2,
\label{eq:DP}
\end{equation}
where $q_f, q_g$ are the respective SRVF representations of $f$, $g$, and $\gamma: [0,1] \rightarrow [0,1]$ is a reparameterization function. The $\gamma$ obtained in this fashion acts as a suitable diffeomorphism between $q_f$ and $q_g$, which in turn produces a warp from $f$ to $g$ in the original space. In fact, we need only recover $\dot{\gamma}$, since $\gamma$ can be reconstructed via 
\begin{equation}
    \gamma(s) = \int_{[0,s]} \dot{\gamma}(t) dt.
    \label{eq:gamma_recon}
\end{equation}

In the general setting, we have a collection of curves $\fn$ where we must make a justified choice of template. In this setting, the K{\"a}rcher mean~\cite{karcher77} is
a well-known average
representation of the overall shape variability. Hence, the K{\"a}rcher
mean, denoted $q_\mu$, serves as a suitable template candidate and is given by
\begin{equation}
q_\mu=\argmin_{q} \frac{1} {N} \sum_{i=1}^{N}  \argmin_{\gamma} d(q - \sqrt{\dot{\gamma}} (q_i \circ \gamma))^2,
\label{eq:karcher}
\end{equation}
where $d(,)$ is the geodesic distance given by, $d(\psi_i, \psi_j)
 = \cos^{-1}\inner{\psi_i}{\psi_j}$, where $\inner{\cdot}{\cdot}$ is the standard $\ltwo$ inner product.

\subsection{Unsupervised Prediction of Warping Functions Under a Fixed Template}\label{sec:fixed_templte}

 In this paper, we minimize a loss function that uses the chord distance $\norm{ q - \sqrt{\dot{\gamma}} (q_i \circ \gamma)}$ instead of the geodesic distance and apply deep learning to learn the desired template
and to learn the proper warping functions without prior knowledge of the warps nor
the K{\"a}rcher mean, effectively making this approach unsupervised. Formally, since the geodesic paths under the SRVF framework converge to a locally unique solution, the optimization problem in (\ref{eq:DP}) produces a unique diffeomorphism, $\gamma$. However, in the unsupervised approach, we don't have prior information of the uniqueness of geodesic paths, so we specify the constraints $\gamma(0) = 0$, $\gamma(1) = 1$, and a non-decreasing condition on $\gamma$. This ensures that $\gamma$ produced by the network is unique.

We start by assuming we have a template $\bar{q}$ and a collection 
$\mathcal{Q} \equiv \{q_i\}_{i=1}^N$
which we would like to warp to $\bar{q}$. As such, we seek to obtain 
$\{\gamma_i\}_{i=1}^N$ that solve equation (\ref{eq:DP}). 
We assume that all functions $\bar{q}$, $\{q_i\}_{i=1}^N$ have been discretized
into $T$ uniformly spaced points over the domain $[0,1]$. 
We construct an 
unsupervised deep learning framework based on the variational autoencoder \cite{kingma2013} to 
subsequently not only obtain the $\gamma_i$ diffeomorphisms, but also an estimate of the distribution of
$\gamma_i$'s, which we denote $\Gamma$, that warp functions from the class $\mathcal{Q}$ to $\bar{q}$.

Adopting the standard autoencoder terminology, our SrvfNet architecure consists of an \emph{encoder} and a \emph{decoder}.
We let $\phi$ denote the trainable parameters of the encoder network, and $\theta$ the parameters of the
decoder. The encoder takes a function $q$ and maps it to an $\ell$-dimensional space via the
reparameterization trick. In particular, the encoder outputs a mean $\mu^\phi(q) \in \Rn{\ell}$ and diagonal 
covariance matrix $\Sigma^\phi(q) \in \real^{\ell \times \ell}$ which are then used to construct the 
low-dimensional representation $z \in \Rn{\ell}$ defined by 

\begin{align}\label{eq:reparm_trick}
z = \left(\Sigma^\phi(q)\right)^\frac{1}{2} \tilde{z} + \mu^\phi(q)
\end{align}
where $\tilde{z} \sim \mathcal{N}(0, I_{\ell})$. This has the effect of modeling the posterior 
distribution of the latent variable $z$ given $q$ as 
$p^\phi(z|q) = \mathcal{N}(\mu^\phi(q), \Sigma^\phi(q))$ which is then transformed into a distribution
over the diffeomorphisms by the decoder. We also invoke a prior on the latent variables
$z \sim \mathcal{N}(0,I_{\ell})$ to allow for efficient sampling over $\Gamma$. More specifically,
once our network is trained, we can generate a sample from $\Gamma$ by sampling 
$z \sim \mathcal{N}(0,I_{\ell})$ and then pass $z$ through the decoder. This gives rise to the first
term in our training loss function where we push the posterior $p^\phi(z|q)$ to resemble the prior 
$p(z)$ and measure this disparity through the Kullback-Leibler (KL) divergence. Because the posterior
and prior distributions are assumed to be normal, the KL divergence exhibits a closed-form solution.
Letting $\Lcal_{KL}(q) = \KL{p^\phi(z|q)}{p(z)}$, we have

\begin{align}
\nonumber
\Lcal_{KL}(q) = \frac{1}{2} \bigl[
\trace{\Sigma^\phi(q)} - l &+ \left(\mu^\phi(q)\right)^T\mu^\phi(q) \\
\label{eq:KL_loss}
&- 
\log \left| \Sigma^\phi(q) \right|\bigr].
\end{align}
Once $z$ is obtained using equation (\ref{eq:reparm_trick}), it is propagated  through the decoder 
which outputs $v^{\theta} \in \Rn{T}$ which then goes through a diffeomorphic constraint
satisfaction layer. We construct $\gamma$ by first transforming $v^\theta$ into an estimate
of $\dot{\gamma}$ which we denote $\dot{\gamma}^\theta$. Similar to Lohit et al.~\cite{lohit_transformer}, we first map $v^{\theta}$ onto the probability simplex by normalizing and then take the Hadamard product
of the resulting output as given by

\begin{align}\label{eq:decoder_gamma_dot}
\dot{\gamma}^\theta &= \frac{v^\theta}{\norm{v^\theta}} \odot 
\frac{v^\theta}{\norm{v^\theta}}.
\end{align}
In conjunction with equation (\ref{eq:gamma_recon}), the estimated $\gamma$, denoted $\gamma^\theta$, is then 
constructed as 

\begin{align}\label{eq:decoder_gamma}
\gamma^\theta(s) &= \sum_{t=1}^s \dot{\gamma}^\theta (t).
\end{align}
Normalizing $v^\theta$ by $\norm{v^\theta}$ in equation (\ref{eq:decoder_gamma_dot})
ensures that all elements in 
$\frac{v^\theta}{\norm{v^\theta}}$ are in the interval $[-1,1]$, and taking the Hadamard product
ensures all elements of $\dot{\gamma}^\theta$ are nonnegative, which leads to a non-decreasing
$\gamma^\theta$.

We further impose the constraint that $\gamma^\theta(0) = 0$ and $\gamma^\theta(1) = 1.$
To promote smooth diffeomorphisms, we use linear interpolation to downsample 
$\gamma^\theta \in \Rn{T}$ to $\tilde{\gamma}^\theta \in \Rn{\tilde{T}}$ uniformly
on $[0,1]$ where $\tilde{T}  < T$. We then use linear interpolation again to upsample 
$\tilde{\gamma}^\theta$ back to
$\Rn{T}$. With a slight abuse of notation, we refer to the upsampled $\tilde{\gamma}^\theta$ 
as $\gamma^\theta$.

The second term in our training loss function is inspired by the Fisher-Rao metric defined in 
equation (\ref{eq:DP}) and is given by

\begin{align}\label{eq:fr_fixed}
    \Lcal_{FR}(q, \gamma^\theta; \bar{q}) &= 
    \norm{\bar{q} - \sqrt{\dot{\gamma}^\theta} (q \circ \gamma^\theta)}_2^2.
\end{align}

To penalize diffeomorphisms with large slopes and to further promote smoothness, we consider 
penalties on the norms of both the first and second derivatives of $\gamma^\theta$. For the first
derivative, we define

\begin{align}\label{eq:first_derivative_penalty}
\Lcal_\nabla \left(\gamma^\theta\right) &= \norm{\dot{\gamma}^\theta}_2^2,
\end{align}
and for the second derivative,

\begin{align}\label{eq:second_derivative_penalty}
\Lcal_{\nabla^2} \left(\gamma^\theta\right) &= \norm{\ddot{\gamma}^\theta}_2^2.
\end{align}

The loss function we consider is a weighted sum of 
$\Lcal_{KL}, \Lcal_{FR}, \Lcal_{\nabla}$, and $\Lcal_{\nabla^2}$ given by
equations (\ref{eq:KL_loss}),
(\ref{eq:fr_fixed}), (\ref{eq:first_derivative_penalty}), and (\ref{eq:second_derivative_penalty})
with respective weights $\lambda_{KL}, \lambda_{FR}, \lambda_{\nabla}, \lambda_{\nabla^2}$.
For a fixed template $\bar{q}$ and function $q$ with predicted warping function
$\gamma^\theta$, the loss is therefore given by

\begin{align}
\nonumber
\Lcal_{F} = \lambda_{FR} \Lcal_{FR}\left(q, \gamma^\theta; \bar{q}\right) 
+ \lambda_{KL}\Lcal_{KL}(q)
&+ \lambda_\nabla \Lcal_\nabla \left(\gamma^\theta\right) \\
\label{eq:fixed_template_loss}
&+ \lambda_{\nabla^2} \Lcal_{\nabla^2} \left(\gamma^\theta\right)
\end{align}
For a training batch $\{q_i\}_{i=1}^B$, we define the loss as 

\begin{align}
\nonumber
\Lcal_{F} = \frac{1}{B}\sum_{i=1}^B \bigl[&\lambda_{FR} \Lcal_{FR}\left(q_i, \gamma_i^\theta; \bar{q}\right) 
+ \lambda_{KL}\Lcal_{KL}(q_i) \\
\label{eq:fixed_template_batch_loss}
&+ \lambda_\nabla \Lcal_\nabla \left(\gamma_i^\theta\right) 
+ \lambda_{\nabla^2} \Lcal_{\nabla^2} \left(\gamma_i^\theta\right)\bigr].
\end{align}

\subsection{Unsupervised Joint Prediction of the Template and Warping Functions}\label{sec:est_template}

In the previous section we considered the case where a template $\bar{q}$ was provided
to us a priori. This could be, for example, the K{\"a}rcher mean of the collection 
$\mathcal{Q}$ as defined by equation (\ref{eq:karcher}), or it could be some $q_i \in 
\mathcal{Q}$, or any other function in $\mathcal{S}$. In this section we will consider
the case where the template is unknown to us, and we wish to estimate the optimal template 
to warp the collection $\mathcal{Q}$ to. We would like this predicted template to capture the
geometric variability of $\mathcal{Q}$, and as such aim to integrate the notion of the  K{\"a}rcher mean into
our previous Fisher-Rao loss defined in equation (\ref{eq:fr_fixed}). In particular, 
for batch $\{q_i\}_{i=1}^B$, we estimate the template as 

\begin{align}\label{eq:template_est_def}
\hat{q} &= \Omega\left(\frac{1}{B} \sum_{i=1}^B \sqrt{\dot{\gamma}_i^\theta} 
\left(q_i \circ \gamma_i^\theta\right)\right)
\end{align}
where $\Omega(u) = \frac{u}{||u||_2}$ normalizes its input. In other words, we estimate the
template as the Euclidean mean of the warped batch and then normalize to ensure
$\hat{q} \in \mathcal{S}$.

The batch template $\hat{q}$ can now be used as a surrogate for the template
$\bar{q}$ as given in equation (\ref{eq:fr_fixed}). For a given batch, our Fisher-Rao loss 
function takes the form

\begin{align}\label{eq:fr_est}
\hat{\Lcal}_{FR}(q, \gamma^{\theta}; \hat{q}) &=
\norm{\hat{q} - \sqrt{\dot{\gamma}^\theta} (q \circ \gamma^\theta)}_2^2
\end{align}
where $\hat{q}$ is as defined in equation (\ref{eq:template_est_def}).
The remaining terms in our loss function mirror those from the fixed template setting in equation
(\ref{eq:fixed_template_batch_loss}). More precisely, we will again have weights 
$\hat{\lambda}_{FR}, \lambda_{KL},  \lambda_{\nabla}, \lambda_{\nabla^2}$ for the 
Fisher-Rao loss with template prediction, KL loss, and first and second derivative penalties
as defined in equations (\ref{eq:fr_est}), (\ref{eq:KL_loss}),
(\ref{eq:first_derivative_penalty}), and (\ref{eq:second_derivative_penalty}), respectivly.
In this setting, we aim to minimize the following loss defined for batch $\{q_i\}_{i=1}^B$

\begin{align}
\nonumber
\Lcal_{E} = \frac{1}{B}\sum_{i=1}^B \bigl[&\hat{\lambda}_{FR} 
\hat{\Lcal}_{FR}\left(q_i, \gamma_i^\theta; \hat{q}\right) 
+ \lambda_{KL}\Lcal_{KL}(q_i) \\
\label{eq:est_template_batch_loss}
&+ \lambda_\nabla \Lcal_\nabla \left(\gamma_i^\theta\right) 
+ \lambda_{\nabla^2} \Lcal_{\nabla^2} \left(\gamma_i^\theta\right)\bigr].
\end{align}

While our setup does not explicitly output a predicted template, once our network has been trained
we can obtain an estimate of the predicted template by computing equation (\ref{eq:template_est_def})
over the entire training data. 
\begin{figure*}[htb]
\begin{center}
\includegraphics[width=\linewidth]{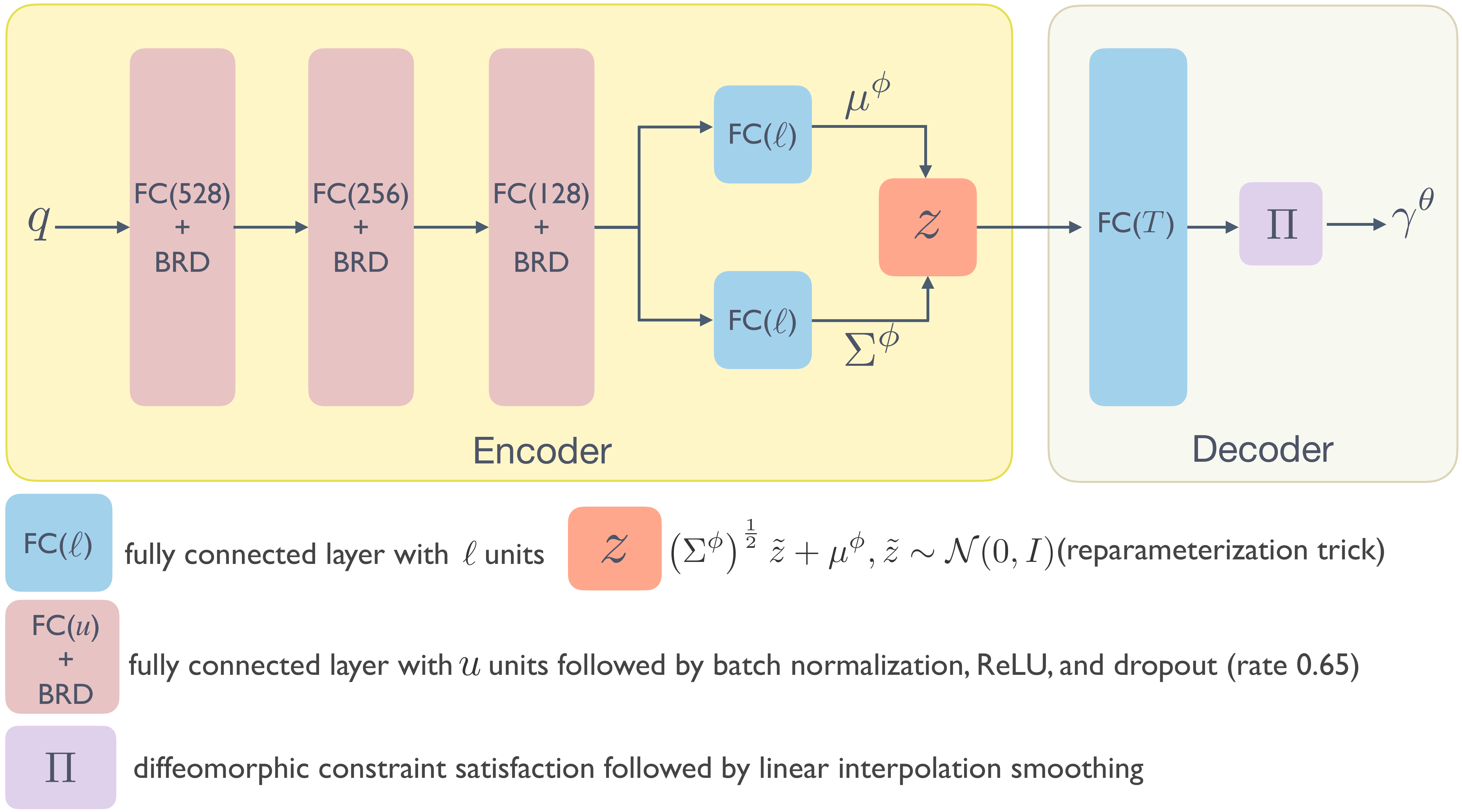}
\end{center}
\caption{Schematic of the SrvfNet architecture.}
\label{fig:network_arch}
\end{figure*}
\section{Network Design}\label{sec:network_design}

The SrvfNet architecture is designed to operate on both the fixed template setting 
described in section \ref{sec:fixed_templte} and the template prediction setting from
section \ref{sec:est_template}. The training regimes differ only in
their respective loss functions given by equations (\ref{eq:fixed_template_batch_loss}) and
(\ref{eq:est_template_batch_loss}). 

\subsection{SrvfNet Architecture}

A schematic diagram of the SrvfNet architecture is provided in figure \ref{fig:network_arch}. Both our encoder
and decoder networks consist only of fully-connected layers. We let BRD represent a batch 
normalization layer, followed by a ReLU activation, and finally a dropout layer with drop rate 0.65.
Our encoder, then, consists of three fully-connected-BRD pairs with 528, 256, and 128 
fully-connected units, respectively.
The output of this last layer is then passed to two fully-connected layers, each
with $\ell$ units. The first represents the mean $\mu^{\phi}(q)$, while the second represents the
diagonal covariance matrix $\Sigma^\phi(q)$ (for numerical stability, this layer outputs 
$\log\left(\Sigma^\phi(q)\right)$ which is exponentiated to recover $\Sigma^\phi(q)$). 
$\mu^\phi(q)$ and $\Sigma^\phi(q)$ are then used to sample 
$z \sim \mathcal{N}\left(\mu^\phi(q), \Sigma^\phi(q)\right)$ as in equation (\ref{eq:reparm_trick}).

Given $z$, we obtain $\gamma^\theta$ by passing $z$ through the decoder network. Our decoder consists
of a single fully-connected layer with $T$ units which is then followed by a diffeomorphic constraint
satisfaction and smoothing layer, which we denote $\Pi$, as discussed in section 
\ref{sec:fixed_templte}.

\subsection{Training and Implementation Details}

All network weights are initialized using a Glorot Uniform initialization 
\cite{glorot2010understanding}. We use a batch size of 512 and an Adam optimizer 
\cite{kingma2015} with learning rate $10^{-3}$. Loss function weights as defined in 
equations (\ref{eq:fixed_template_batch_loss}) and (\ref{eq:est_template_batch_loss}),
as well as number of training epochs, are dataset-dependent and discussed further in subsequent 
sections. All models are implemented using TensorFlow on   an Intel i7-7700K CPU @ 4.20GHz machine equipped with TITAN Xp GPUs.



\section{Experiments}\label{sec:experiments}
\subsection{Datasets}

\begin{figure*}[htb]
\begin{center}
\includegraphics[width=\linewidth]{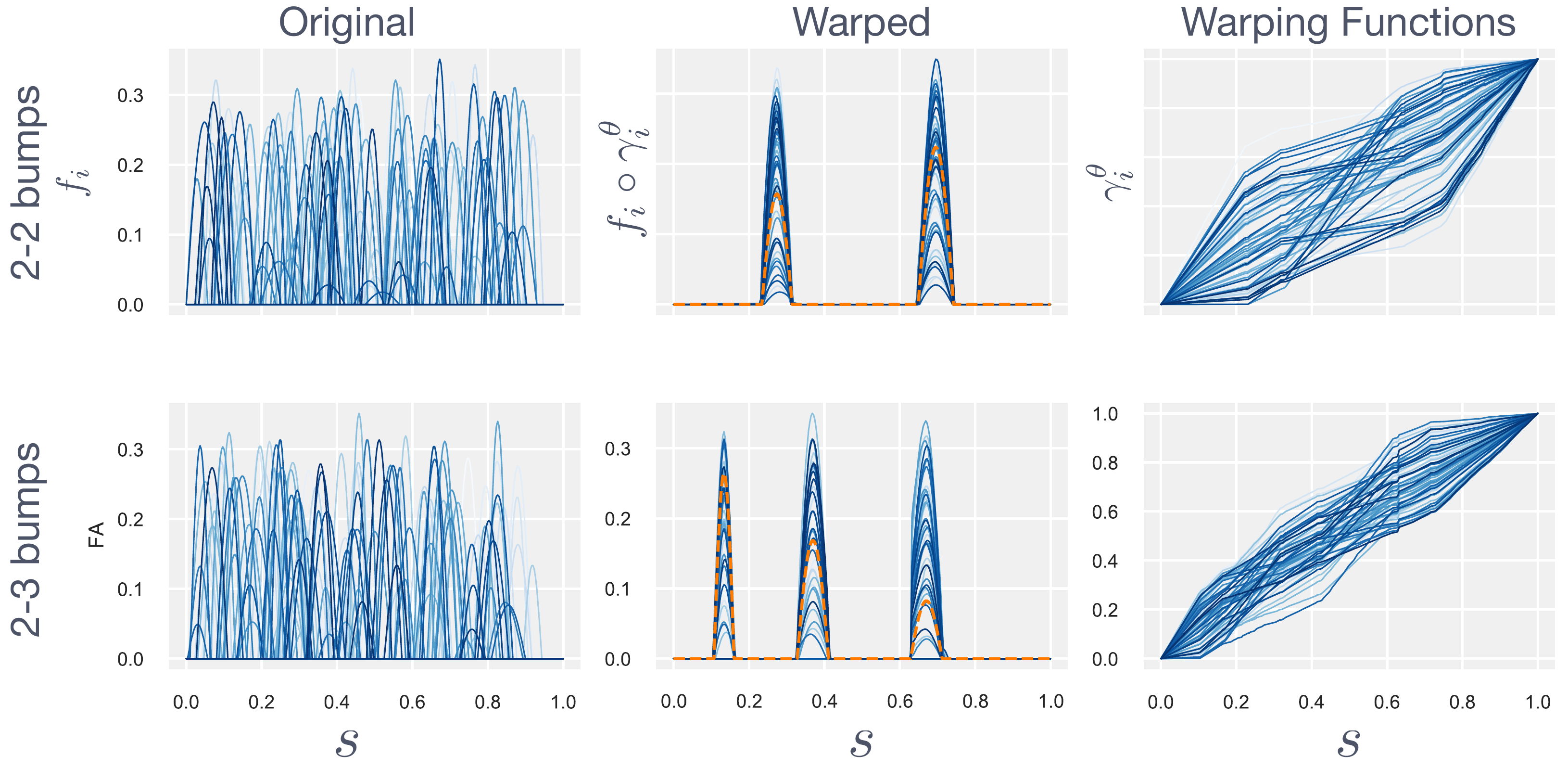}
\end{center}
\caption{Two-to-Two and Two-to-Three bump matching with template (in orange).}
\label{fig:fixed_bumps}
\end{figure*}

Our data comprises two classes i) synthetic bump functions, and ii) diffusion profiles from MRI data.  
We first consider bump functions as shown in the first (original) column of 
figure \ref{fig:fixed_bumps}. In this data we consider a collection of 
randomly generated sinusoidal functions characterized by their number
of peaks, amplitudes, and phases. Each function is discretized to $T = 300$ 
uniformly-spaced points on the interval $[0,1]$.
We consider a two-bump dataset that consists
of bump functions with two peaks and a three-bump dataset that consists of
bump functions with three peaks. 

We also consider a more realistic dataset consisting of 
fractional anisotropy (FA) values derived from white matter fiber bundles as depicted in the `original'
column of figure \ref{fig:est_bundles}. In this dataset, each function is discretized
to $T = 100$ uniformly-spaced points on the interval $[0,1]$.
We consider five bundle subclasses: i) 
the arcuate tract, ii) corpus callosum forceps minor (CCFmin), iii) cortico spinal tract (CST), iv) 
superior longitudinal fasciculus (SLF), and v) thalamic radiation tract (Th Rad). 

\subsection{Unsupervised Warping to a Fixed Template}

In the fixed template setting, we perform two experiments on the bump datasets. 

\subsubsection{Two-to-Two Bumps Matching}\label{sec:two_to_two_fixed_exp}
We first generate a random bump function with two peaks and designate this as the template.
We then randomly generate $50,000$ bump functions to serve as our training dataset and
$5,000$ more to serve as our test set. All bump functions are converted to their SRVF
representations and are normalized to have unit length. We then train our network to minimize
the fixed-template loss function given by equation (\ref{eq:fixed_template_batch_loss}).
We train for 5000 epochs (approximately 30 minutes of training) and use a latent space
dimension of $\ell = 150$.
After training, we 
sample 70 bumps from our test set and obtain the predicted warping functions and apply
them to the sampled functions. Figure
\ref{fig:fixed_bumps} shows the test data, the warped data to a fixed template (orange) and the predicted warping functions ${\gamma_i^{\theta}}$.

\subsubsection{Two-to-Three Bumps Matching}
\begin{figure}[htb]
\begin{center}
\includegraphics[width=0.5\textwidth]{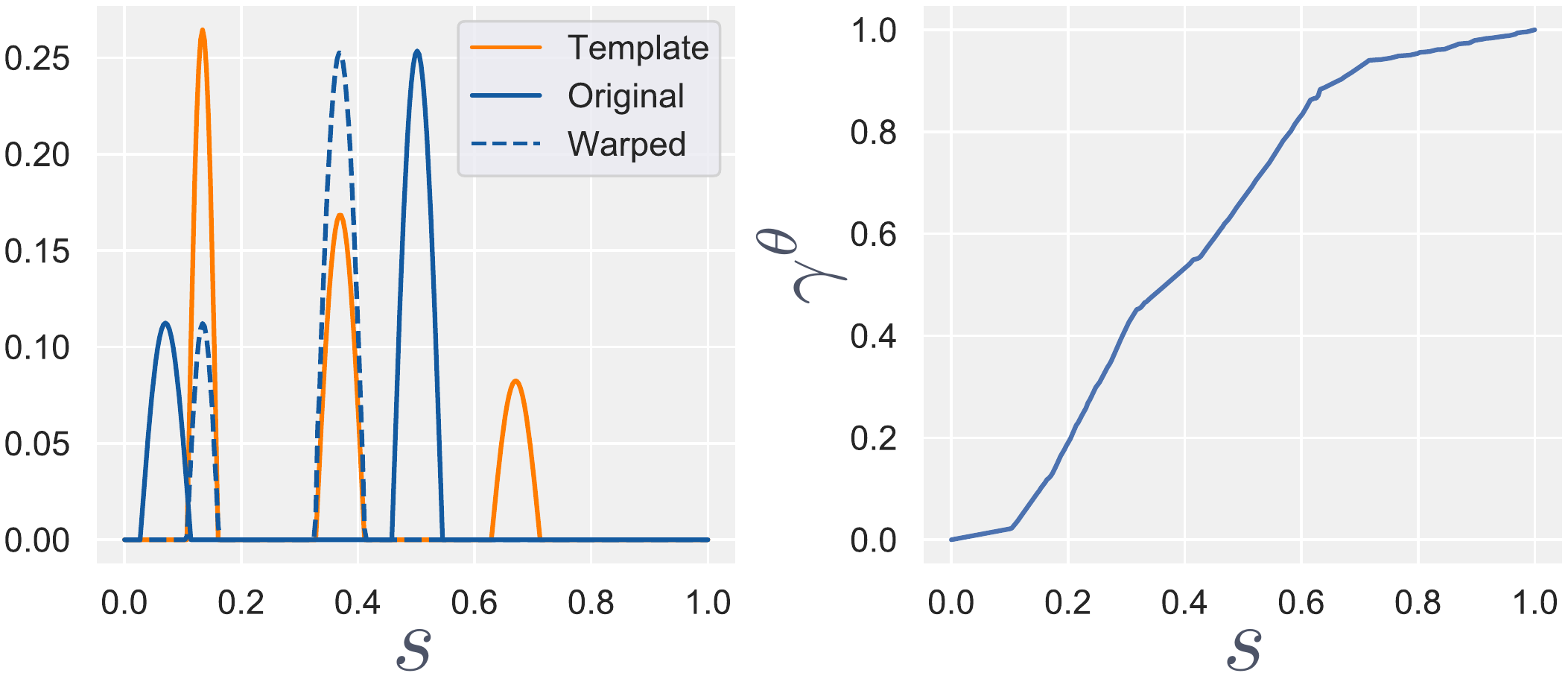}
\end{center}
\caption{Two-to-Three bump matching with template (in orange).}
\label{fig:fixed_bump_single}
\end{figure}

In this experiment we reuse the two-bump training and test sets constructed in the previous 
experiment, but replace the template with a bump function that has three peaks.  We train for 5000 epochs with a $150$-dimensional latent space  and visualize the performance on 
$70$ randomly-sampled
test functions as shown in the second row of figure \ref{fig:fixed_bumps}. To clarify the nature of the warping functions learned by our network, we visualize the process on a single
function in figure \ref{fig:fixed_bump_single}, where we see two peaks in the test function warp to two out of three peaks in the template (orange).

\subsubsection{Random Samples in the Encoding space of Warps}
\begin{figure}[htb]
\begin{center}
\includegraphics[width=\linewidth]{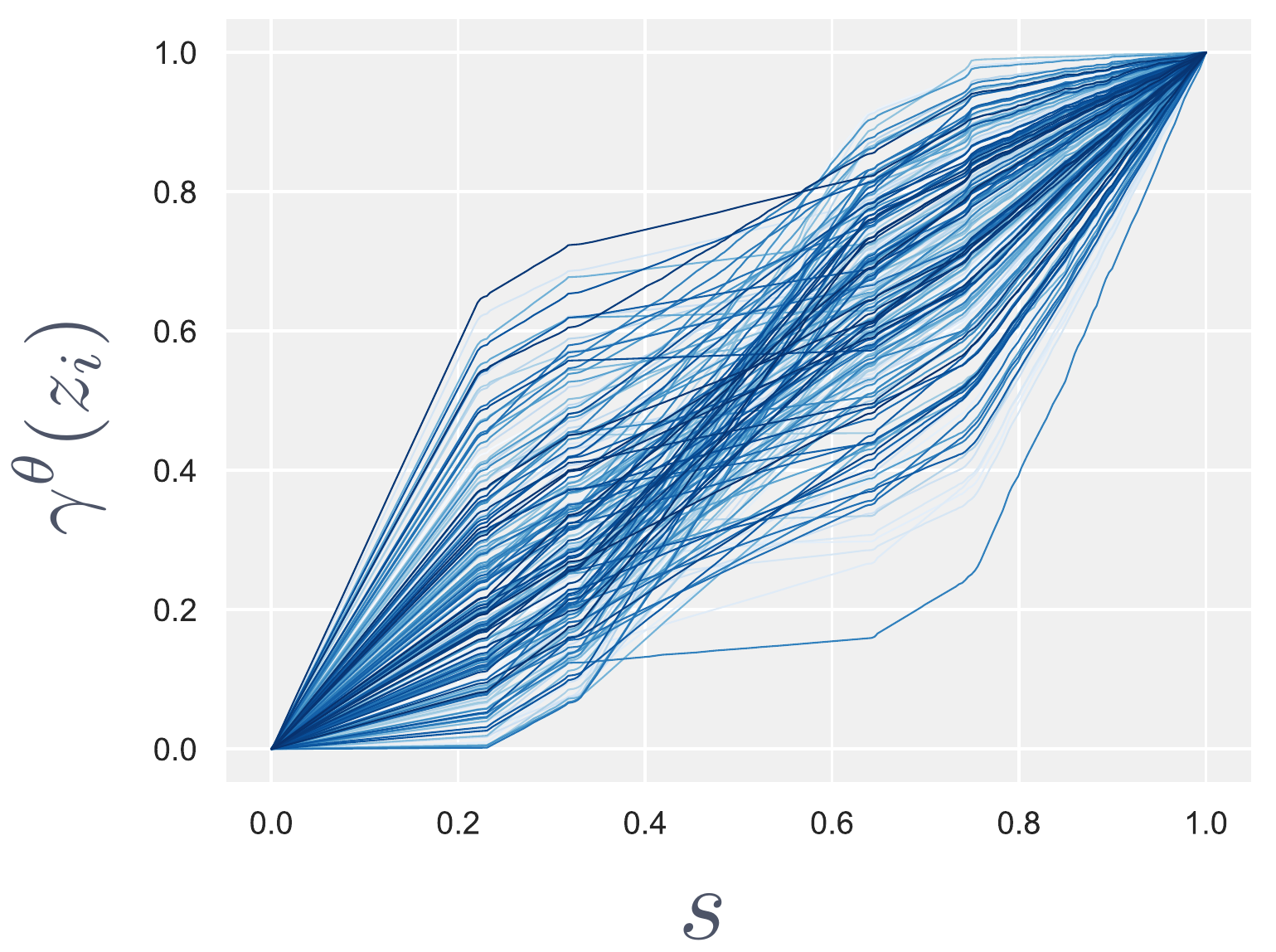}
\end{center}
\caption{Distribution of $\gamma$ for two-to-two bump matching with a fixed template.}
\label{fig:fixed_two_two_gamma_dist}
\end{figure}

As discussed in section \ref{sec:fixed_templte}, our network allows sampling from the estimated space of 
diffeomorphisms that warp the training functions to the prescribed template by repeatedly sampling 
$z_i \sim \mathcal{N}(0,I_{\ell})$ and passing this through the decoder. To qualitatively demonstrate that this procedure yields valid warping functions, we  randomly sampled $200$ $z_i$'s and passed them through the decoder for the two-bump template model to obtain warping functions $\gamma^\theta(z_i)$. Figure  \ref{fig:fixed_two_two_gamma_dist} visualizes these results, which empirically shows the distribution over warps that  encodes the intrinsic relationships between the class of two-bump functions and the template.

\subsection{Unsupervised Template Prediction}

In the template prediction setting, we show results on synthetic data and diffusion profile data from fiber bundles.

\subsubsection{Two-to-Two Bumps}

\begin{figure*}[htb]
\begin{center}
\includegraphics[width=\linewidth]{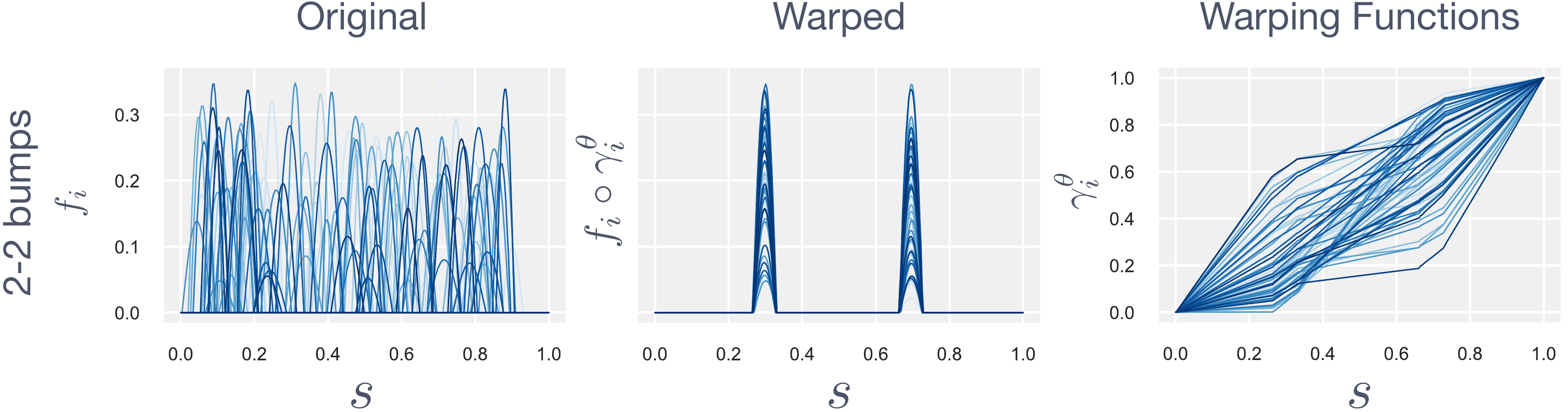}
\end{center}
\caption{Warping of Two-to-Two bump functions with simultaneous template prediction.}
\label{fig:est_bumps}
\end{figure*}

Here we reuse the train and test datasets constructed for the two-bump dataset from
section \ref{sec:two_to_two_fixed_exp}. However, instead of using a template $\bar{q}$, we minimize the loss function given in equation 
(\ref{eq:est_template_batch_loss})
and learn an appropriate template from the training data. We  train for 1200 epochs 
with $\ell = 150$ and
visualize the performance of our trained model on $70$ randomly sampled test functions in figure 
\ref{fig:est_bumps}.

\subsubsection{Fractional Anisotropy (FA) Profiles}

\begin{figure*}[htb]
\begin{center}
\includegraphics[width=\linewidth]{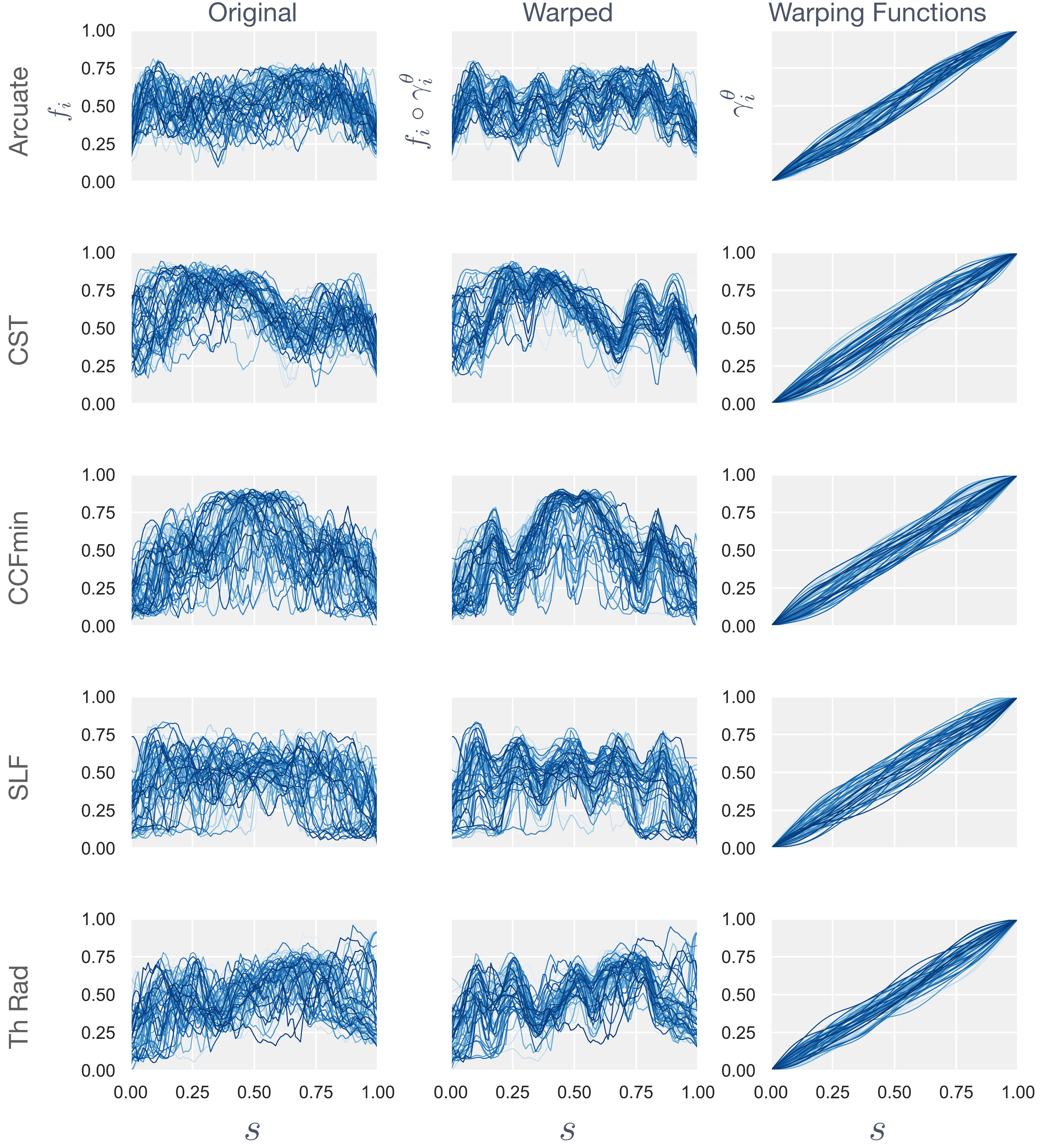}
\end{center}
\caption{Warping of FA diffusion profiles with simultaneous template prediction.}
\label{fig:est_bundles}
\end{figure*}

We train five different models on each of the bundles. Our training sets for 
Arcuate, CCFmin, CST, SLF, and Th Rad, have sizes $300$K, $260$K, $110$K, $390$K, and $160$K, 
respectively. Test sets ranged in size from $20$K to $90$K. 
We again aim to minimize the loss function given in equation (\ref{eq:est_template_batch_loss})
and train for $1250$ epochs (about $15-30$ minutes of training) with $\ell = 50$.
We visualize the performance
of our models on 70 randomly sampled functions from our test sets in figure \ref{fig:est_bundles}.
After the warping phase, one can see more pronounced patterns of shapes in the underlying FA values.



\section{Discussion}\label{sec:discussion}
\subsection{Fixed Template}
From figure \ref{fig:fixed_bumps} we observe that SrvfNet can successfully warp the functions in the test set to the template. Additionally, the produced warping functions appear as piecewise affine warps as expected when mapping a two-bump
function to another bump function. We also replicated this experiment using the standard DTW method and found that the procedure took 
approximately 11 minutes to warp $50,000$ two-bump functions to a three-bump template on a 2.6GHz 6‑core Intel Core i7 processor.  In contrast, the training phase of our network is slower and takes approximately 30 minutes to warp the training set to the prescribed  template. While training time is slower, once the network is trained, the 
efficient implementation 
of the forward pass allows for extremely fast warpings of unseen functions belonging to the same class of  training data.

\subsection{Template Prediction}

SrvfNet provides a convenient way to compute a population template by computing a mean of the set of warped inputs. Since all
functions will be nonlinearly aligned  to this underlying predicted template, their mean will resemble the predicted template. By simply visualizing the warped inputs as  in figure  
\ref{fig:est_bumps} for the two-to-two bump experiment, we can also gain an insight into the types of functions that the model learns for the purpose of warping. For this two-to-two bump experiment, the network 
was able to learn a suitable two-bump template function, as evidenced by the two peaks in the warped outputs. Moreover,
the predicted warping functions continue to exhibit the piecewise affine structure that we
expect and observed in the fixed template case. We also note that while our network may have 
learned a template that minimized equation (\ref{eq:est_template_batch_loss}), this 
solution may not be unique as there may be several functions that yield the same objective.
Indeed, in our experiments we found that different initializations lead to varying predicted templates with similar loss function values. To replicate this experiment using DTW we first need to construct a template as there is not one 
provided in this setting. Since our framework aims to construct a template that captures the geometric properties of the data,
we would like our DTW template to also encode the shape variability of the training data. 
As discussed in \ref{sec:shape_rec}, one such suitable template is the K{\"a}rcher mean in equation (\ref{eq:karcher}), from which our loss function (\ref{eq:est_template_batch_loss}) is based. 
However, solving equation 
(\ref{eq:karcher}) becomes prohibitively expensive for large datasets. As an example, solving
(\ref{eq:karcher}) for $100K$  diffusion profiles would take roughly $25$ hours on our platform. This is $150$ times  slower than our approach, which takes less than $30$ minutes to train on $300K$  diffusion profiles, while also producing a template that encodes the shape variability of the profiles.

%


In general, we have no guarantees on the  uniqueness of the predicted template. In all our experiments we assumed that the signal-to-noise ratio (SNR) is acceptable ($> 15$ dB after following standard processing techniques for MRI data).  Further, in the case of functions and signals, where the noise levels are comparable to that of signal amplitudes, there may be ambiguity in generating the warping functions in the training phase. This may give rise to spurious peaks or valleys, which may not naturally occur in the training data. While we did not observe this behaviour in our experiments with bump functions, data generated from biological processes such as diffusion profiles may consist of segments where the signal to noise ratio may be reduced. Thus more validation of the method is needed in cases where such SNR assumptions are violated. 

\section{Conclusion}\label{sec:conclusion}

The proposed  generative encoder-decoder network SrvfNet  is  capable of efficiently computing 
warping functions in an unsupervised manner that is also amenable to both
fixed template and template prediction schemes. We validated our fixed template 
models on two experiments and found that the resulting warpings strongly resembled what we would expect 
from standard dynamic time warping approaches. Similarly, we validated our template prediction
framework on both synthetic and FA data. In the synthesized bumps, we found that our predicted
template closely resembled an exemplar function from the class of training data and displayed a prominent feature of two bumps.
This suggests that our framework has learned the geometric structure of the data and has produced
a more descriptive template than a simple Euclidean mean, which would not necessarily preserve the geometric
properties of the data. This is also demonstrated in the FA experiments where we observed  distinct patterns in diffusion profiles  in the warped test data compared to the original profiles  while still retaining the overall structure of the original data.
Importantly, our approach is not only capable of jointly aligning multiple signals and estimating a nonlinear template, but it also allows for sampling of warping functions from a space of distributions of encoded warps. 

\section {Acknowledgements}
This research was partially supported by a fellowship from the NSF NRT Award \#$1829071$ (EN) and the 
NIH NIAAA (National Institute on Alcohol Abuse and Alcoholism)  awards R01-$AA026834$, R01-$AA025653$ (SHJ). We also thank Dr. Katherine L. Narr, UCLA, for providing the diffusion profile data. 

\bibliographystyle{IEEEtran}
\bibliography{paper}

\begin{thebibliography}{10}
\providecommand{\url}[1]{#1}
\csname url@samestyle\endcsname
\providecommand{\newblock}{\relax}
\providecommand{\bibinfo}[2]{#2}
\providecommand{\BIBentrySTDinterwordspacing}{\spaceskip=0pt\relax}
\providecommand{\BIBentryALTinterwordstretchfactor}{4}
\providecommand{\BIBentryALTinterwordspacing}{\spaceskip=\fontdimen2\font plus
\BIBentryALTinterwordstretchfactor\fontdimen3\font minus
  \fontdimen4\font\relax}
\providecommand{\BIBforeignlanguage}[2]{{%
\expandafter\ifx\csname l@#1\endcsname\relax
\typeout{** WARNING: IEEEtran.bst: No hyphenation pattern has been}%
\typeout{** loaded for the language `#1'. Using the pattern for}%
\typeout{** the default language instead.}%
\else
\language=\csname l@#1\endcsname
\fi
#2}}
\providecommand{\BIBdecl}{\relax}
\BIBdecl

\bibitem{srivastava2016functional}
A.~Srivastava and E.~P. Klassen, \emph{Functional and Shape Data
  Analysis}.\hskip 1em plus 0.5em minus 0.4em\relax Springer, 2016, vol.~1.

\bibitem{sakoe1978dynamic}
H.~Sakoe and S.~Chiba, ``Dynamic programming algorithm optimization for spoken
  word recognition,'' \emph{IEEE Trans. on Acoustics, Speech, and Signal
  Processing}, vol.~26, no.~1, pp. 43--49, 1978.

\bibitem{kazlauskaite2018gaussian}
I.~Kazlauskaite, C.~H. Ek, and N.~D. Campbell, ``Gaussian process latent
  variable alignment learning,'' \emph{arXiv preprint arXiv:1803.02603}, 2018.

\bibitem{trigeorgis2017deep}
G.~Trigeorgis, M.~A. Nicolaou, B.~W. Schuller, and S.~Zafeiriou, ``Deep
  canonical time warping for simultaneous alignment and representation learning
  of sequences,'' \emph{IEEE Trans. on Pattern Analysis and Machine
  Intelligence}, vol.~40, no.~5, pp. 1128--1138, 2017.

\bibitem{Oh_invariance:2018}
J.~Oh, J.~Wang, and J.~Wiens, ``Learning to exploit invariances in clinical
  time-series data using sequence transformer networks,'' in \emph{Proceedings
  of the 3rd Machine Learning for Healthcare Conference}, vol.~85, 2018, pp.
  332--347.

\bibitem{lohit_transformer}
S.~Lohit, Q.~Wang, and P.~Turaga, ``Temporal transformer networks: Joint
  learning of invariant and discriminative time warping,'' in \emph{Proceedings
  of the IEEE/CVF Conference on Computer Vision and Pattern Recognition}, 2019,
  pp. 12\,426--12\,435.

\bibitem{Abid:18}
A.~Abid and J.~Y. Zou, ``Learning a warping distance from unlabeled time series
  using sequence autoencoders,'' \emph{Advances in Neural Information
  Processing Systems}, vol.~31, 2018.

\bibitem{Weber:19}
R.~A. Shapira~Weber, M.~Eyal, N.~Skafte, O.~Shriki, and O.~Freifeld,
  ``Diffeomorphic temporal alignment nets,'' \emph{Advances in Neural
  Information Processing Systems}, 2019.

\bibitem{freifeld2015highly}
O.~Freifeld, S.~Hauberg, K.~Batmanghelich, and J.~W. Fisher,
  ``Highly-expressive spaces of well-behaved transformations: Keeping it
  simple,'' in \emph{Proceedings of the IEEE International Conference on
  Computer Vision}, 2015, pp. 2911--2919.

\bibitem{koneripalli2020rate}
K.~Koneripalli, S.~Lohit, R.~Anirudh, and P.~Turaga, ``Rate-invariant
  autoencoding of time-series,'' in \emph{IEEE International Conference on
  Acoustics, Speech and Signal Processing (ICASSP)}, 2020, pp. 3732--3736.

\bibitem{nunez_CVPRW:20}
E.~Nunez and S.~H.Joshi, ``Deep learning of warping functions for shape
  analysis,'' \emph{2020 IEEE/CVF Conference on Computer Vision and Pattern
  Recognition Workshops (CVPRW)}, pp. 3782--3790, 2020.

\bibitem{S.H.Joshi_etal_CVPR:07}
S.~H. Joshi, E.~Klassen, A.~Srivastava, and I.~Jermyn, ``A novel representation
  for riemannian analysis of elastic curves in ${R}^n$,'' \emph{2007 IEEE
  Conference on Computer Vision and Pattern Recognition}, pp. 1--7, 2007.

\bibitem{kingma2013}
D.~Kingma and M.~Welling, ``Auto-encoding variational bayes,'' \emph{ICLR}, 12
  2013.

\bibitem{Joshi2007emmcvpr}
S.~H. Joshi, E.~Klassen, A.~Srivastava, and I.~Jermyn, ``{Removing
  shape-preserving transformations in square-root elastic (SRE) framework for
  shape analysis of curves},'' in \emph{Energy Minimization Methods in Computer
  Vision and Pattern Recognition (EMMCVPR)}, 2007, pp. 387--398.

\bibitem{srivastava_etal_PAMI:11}
A.~Srivastava, E.~Klassen, S.~H. Joshi, and I.~H. Jermyn, ``Shape analysis of
  elastic curves in {E}uclidean spaces,'' \emph{IEEE Trans. on Pattern Analysis
  and Machine Intelligence}, vol.~33, pp. 1415--1428, 2011.

\bibitem{srivastava_etal_CVPR:07}
A.~Srivastava, I.~H. Jermyn, and S.~H. Joshi, ``Riemannian analysis of
  probability density functions with applications in vision,'' \emph{IEEE
  Conference on Computer Vision and Pattern Recognition}, pp. 1--8, 2007.

\bibitem{karcher77}
H.~Karcher, ``Riemannian center of mass and mollifier smoothing,''
  \emph{Communications on Pure and Applied Mathematics}, vol.~30, pp. 509--541,
  1977.

\bibitem{glorot2010understanding}
X.~Glorot and Y.~Bengio, ``Understanding the difficulty of training deep
  feedforward neural networks,'' in \emph{Proceedings of the Thirteenth
  International Conference on Artificial Intelligence and Statistics}.\hskip
  1em plus 0.5em minus 0.4em\relax JMLR Workshop and Conference Proceedings,
  2010, pp. 249--256.

\bibitem{kingma2015}
D.~P. Kingma and J.~Ba, ``Adam: A method for stochastic optimization,'' in
  \emph{3rd International Conference on Learning Representations}, 2015.

\end{thebibliography}

\end{document}